\documentclass[letterpaper, 10 pt, conference]{ieeeconf}  
 
\IEEEoverridecommandlockouts
\usepackage{cite}
\usepackage{amsmath,amssymb,amsfonts}
\usepackage{algorithmicx}
\usepackage{algorithm}
\usepackage{algpseudocode}
\algrenewcommand\algorithmicrequire{\textbf{Input:}}
\algrenewcommand\algorithmicensure{\textbf{Output:}}
\usepackage{graphicx}
\usepackage{textcomp}
\usepackage{xcolor}
\usepackage{pgfplots}

\usepackage{csquotes}
\usepackage{bbm}
\usepackage{tikz}
\usepackage{pgfplots}
\usepackage{xcolor}
\usepackage{subcaption}
\usepackage{multirow}
\pgfplotsset{compat=1.18}
\usetikzlibrary{positioning}
\usetikzlibrary{decorations.pathreplacing}
\usetikzlibrary{calc}
\usetikzlibrary{positioning,shapes,shadows,arrows}
\usepackage[font=small]{caption}

\pgfplotsset{
every axis/.append style={
  axis line style={->}, 
  legend style={font=\scriptsize},
  label style={font=\scriptsize},
  title style={font=\scriptsize},
  tick label style={font=\scriptsize},
  }
}

\makeatletter
\let\NAT@parse\undefined
\makeatother
\usepackage{hyperref}

\newcommand*{\fix}{\textcolor{black}}

\title{\LARGE \bf ARC-Calib: Autonomous Markerless Camera-to-Robot Calibration via Exploratory Robot Motions
}
    
\author{%
    Podshara Chanrungmaneekul$^{1}$, 
    Yiting Chen$^{1}$,
    Joshua T. Grace$^2$,
    Aaron M. Dollar$^2$,
    Kaiyu Hang$^1$
    \thanks{$^1$Department of Computer Science, Rice University, Houston, TX 77005, USA. $^2$Department of Mechanical Engineering and Material Science, Yale University, New Haven, CT 06511, USA. This work was supported by the US National Science Foundation grant FRR-2133110 and FRR-2132823.}
}

\begin{document}
\maketitle
\begin{abstract}
Camera-to-robot (also known as eye-to-hand) calibration is a critical component of vision-based robot manipulation.
Traditional marker-based methods often require human intervention for system setup. 
Furthermore, existing autonomous markerless calibration methods typically rely on pre-trained robot tracking models that impede their application on edge devices and require fine-tuning for novel robot embodiments.
To address these limitations, this paper proposes a model-based markerless camera-to-robot calibration framework, ARC-Calib, that is fully autonomous and generalizable across diverse robots and scenarios without requiring extensive data collection or learning.
First, exploratory robot motions are introduced to generate easily trackable trajectory-based visual patterns in the camera's image frames.
Then, a geometric optimization framework is proposed to exploit the coplanarity and collinearity constraints from the observed motions to iteratively refine the estimated calibration result.
Our approach eliminates the need for extra effort in either environmental marker setup or data collection and model training, rendering it highly adaptable across a wide range of real-world autonomous systems.
Extensive experiments are conducted in both simulation and the real world to validate its robustness and generalizability.
\end{abstract}

\section{Introduction}
Vision-based robot manipulation tasks rely heavily on accurate camera-to-robot calibration.
The spatial transformation between the robot frame and the camera frame is critical to connecting perception and action in task executions.
Traditional calibration methods \cite{horaud1995hand, park1994robot, ali2019methods, fassi2005hand} are typically formulated as solving the classic  $AX=XB$ equation and requires additional assistance from fiducial markers \cite{olson2011apriltag, fiala2005artag, garrido2014automatic} or chessboards \cite{bennett2014chess}. 
While these methods are well-established and generalizable, they often demand cumbersome manual effort in the system setup processes. 
To eliminate the requirement for manual effort, markerless calibration methods have emerged as a promising alternative \cite{lu2024ctrnet, lee2020camera, lu2023markerless, labbe2021single}.
These approaches develop learning-based frameworks that estimate the camera-to-robot pose directly from environmental observations without marker-aided visual features, thus enabling greater flexibility for real-world autonomous robots.

Despite the progress made, these methods still face non-negligible limitations in generalization.
First, neural networks trained for specific robots are prone to overfit and thus lack the ability to generalize across different unseen embodiments.
The required data collection and policy finetuning for new robots is labor-intensive and impractical at scale.
Second, models trained on synthetic data often require real-world finetuning to overcome the sim-to-real gap.
These limitations motivate research in autonomous markerless camera-to-robot calibration that is both accurate and easily generalizable across diverse robots with unknown geometries and scenarios.

\begin{figure}[!t]   
    \centerline{\input{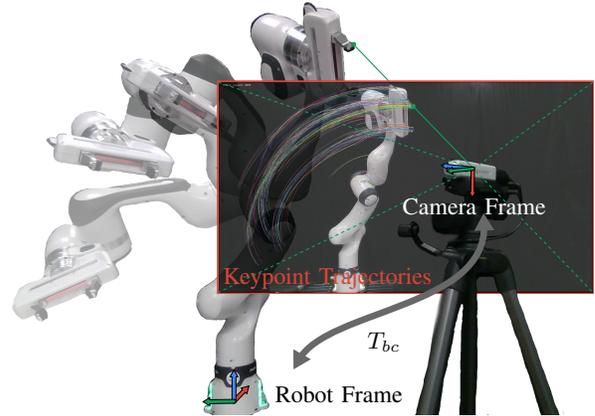}}
    \caption{Our camera-to-robot calibration framework estimates the pose transformation from the camera to the robot via exploratory robot motions. The method relies on analyzing the correspondence between the visual patterns extracted from keypoint trajectories and the robot motion.}
    \label{pic:main_pic}
    \vspace{-0.5cm}
\end{figure}
To address this, we propose ARC-Calib, a model-based autonomous calibration framework that plans exploratory robot motions to generate trackable visual features for iterative camera-to-robot calibration.
As shown in Fig. \ref{pic:main_pic}, instead of
relying on marker-aided or latent space visual features, our key insight lies in examining the correspondence between the structural motion patterns of the robot and the keypoint transformation in the image frame of the camera.
Specifically, our method
consists of
1) a motion planning module that actively selects exploratory motions for the robot to generate corresponding keypoint trajectories in the camera’s image frame and
2) a calibration module that exploits coplanarity and collinearity constraints from the trajectories resulting from motion to optimize the estimated result.
Since the observed keypoint trajectories are motion-oriented and do not contain any semantic information, even traditional image processing algorithms such as optical flow \cite{horn1981determining} are sufficient to extract such visual features for robust calibration.

In summary, the primary contributions of this paper are
listed as follows:
\begin{itemize}
    \item ARC-Calib is a fully autonomous framework that eliminates the need for human intervention, manual setup, or environmental markers, enabling efficient camera-to-robot calibration.
    \item  The calibration framework is generalizable across diverse robots and scenarios without prior database knowledge, pre-trained models, or fine-tuning, promoting broad applicability and adaptability.
    \item ARC-Calib replaces traditional visual markers with exploratory robot motions, generating trackable trajectory-based visual patterns that can be extracted using lightweight methods like optical flow.
\end{itemize}

\section{Related work}
\label{related work}
\textit{Marker-Based Calibration}: 
Robot-to-camera calibration is a fundamental problem in robotic manipulation. 
Traditional approaches include the use of fiducial markers, where early work formulated the hand-eye calibration problem as solving the matrix equation $AX=XB$ on the Euclidean group \cite{horaud1995hand, park1994robot, ali2019methods, fassi2005hand}. 
Chessboard patterns have also been widely adopted given their simplicity, enabling techniques such as simultaneous robot-world-hand-eye calibration \cite{yang2018robotic}. 
Additionally, specialized calibration objects, such as spheres or customized geometries, have been used to facilitate the calibration \cite{tsai1989new, traslosheros2011method} along with methods utilizing RBG-D cameras \cite{staranowicz2015practical}. 
While these methods are accurate, they require manual interventions, tedious setup of markers or objects, and controlled setups, thus limiting their applications in autonomous robot systems.

\textit{Markerless Calibration}:
In contrast, recent markerless calibration approaches leverage advances in deep learning to eliminate the need for additional markers. 
These approaches can be divided into two main types: keypoint-based and rendering-based methods.
Keypoint-based approaches \cite{lu2024ctrnet, lambrecht2019towards, lee2020camera, zuo2019craves} utilize deep neural networks to detect keypoints on the robot and then adapt the perspective-n-point (PnP) algorithm to estimate the calibration results. 
However, inconsistent keypoint detection constrains calibration accuracy. 
Rendering-based methods \cite{labbe2021single, chen2023easyhec} often rely on domain-randomized synthetic training data to enable pose estimation in the real world. 
However, the learned neural network frameworks are trained for specific robot embodiments or scenarios, which require considerable fine-tuning for additional robots.
To address these limitations, our framework utilizes self-exploratory robot motions to create trackable visual patterns, eliminating the need for physical markers or learning-based robot tracking models.

\section{Problem formulation}
\label{problem}
In this work, we address the camera-to-robot calibration problem for $\mathcal{N}$-DoF serial robot manipulators.
We denote the robot's base frame as $T_b \in SE(3)$ and the camera frame as $T_c \in SE(3)$. 
The calibration goal is to estimate the transformation from the robot's base frame to the camera frame, represented as:
\begin{equation}
    T_{bc} = \begin{bmatrix}
    R_{bc} & t_{bc} \\
    \pmb{0} & 1 
    \end{bmatrix}
\end{equation}
where $R_{bc} \in SO(3)$ is the rotation matrix and $t_{bc} \in \mathbb{R}^3$ is the translation vector.
To achieve this, the correspondence between the robot's exploratory actions and trackable visual features is computed.
We find the 3D position of the visual features in the camera frame as follows:
\begin{align}
    s_1 \begin{bmatrix}
    p_x &p_y &1
    \end{bmatrix}^T & = \mathcal{K} \begin{bmatrix}
    u & v & 1
    \end{bmatrix}^T
\end{align}
where $\mathcal{K} $ is the ${3\times3}$ intrinsic matrix of the camera's pinhole projection model, $(p_x, p_y) \in \mathbb{N}^2$ is a pixel coordinate in the image, $s_1\in\mathbb{R}$ is a scaling factor, and $(u,v)\in \mathbb{R}^2$ are Cartesian coordinates obtained by normalizing the pixel coordinates with respect to the intrinsic parameters. 
This normalization ensures that the proposed algorithm remains independent of the camera's intrinsic matrix.
Here, we refer to $(u,v)$ as coordinates on the image plane $\pi$, which can be mapped to 3D points $(s_2u,s_2v,s_2)$ in the homogeneous coordinates of the camera frame for any distance $s_2 \in \mathbb{R}$ in the z-axis. 

To find $T_{bc}$, we propose a markerless calibration framework, which is outlined in Algorithm  \ref{alg:self-calibration}.  
The robot selects an exploratory motion, as detailed in Section \ref{param:motion}, that enables the optical flow algorithm to track keypoint trajectories in the image plane coordinates. 
Using the obtained keypoint trajectories, we estimate the coplanarity and collinearity constraints from the observed trajectory, as explained in Sections \ref{param:rotest} and \ref{param:posest}.
These constraints, corresponding to the robot motion, are then used to estimate the transformation from the robot frame to the camera frame, as described in Section  \ref{calib}. 
By repeating this procedure, the robot iteratively refines the estimated transformation $T_{bc}$ until it converges, as detailed in Section \ref{calib:converge}.
%
\algrenewcommand\algorithmicindent{0.7em}
\begin{algorithm}[t]
    \caption{Markerless Robot-Camera Calibration}
    \label{alg:self-calibration}
    \small
    \begin{algorithmic}[1]
    \Require None
    \Ensure Estimated transformation $T_{bc} \in SE(3)$.
        \State $i\gets0$
        \While {\textbf{not} converged} \hfill \Comment{Sec. \ref{calib:converge}}
            \State $i\gets i+1$
            \State $\mathcal{U}_i \gets \Call{RobotMotionSelection}{ }$ \Comment{Sec. \ref{param:motion}}
            \State $[\mathcal{P}^\pi_{i,1},\dots ,\mathcal{P}^\pi_{i,L_i}] \gets \Call{Execute}{\mathcal{U}_i}$ \Comment{Eq. \eqref{eq:trackpoint}}
            \State $\mathcal{\vec{A}}^c_i,\! \mathcal{\vec{D}}^c_i\!\gets\! \Call{Estimation}{[\mathcal{P}^\pi_{i,1},\dots ,\mathcal{P}^\pi_{i,L_i}]}$ \Comment{Sec. \ref{param:rotest}, \ref{param:posest}}
            \State $T_{bc} \gets \Call{Calibration}{\mathcal{\vec{A}}^c_1, \mathcal{\vec{D}}^c_1, \dots,\mathcal{\vec{A}}^c_i, \mathcal{\vec{D}}^c_i}$ \Comment{Sec. \ref{calib}}
        \EndWhile
        \State \Return $T_{bc}$
    \end{algorithmic} 
\end{algorithm}

\section{Visual Patterns of Exploratory Robot Motions}
\label{param}
In this section, we define the characteristics of exploratory motions.
As the robot moves, its features are tracked, producing a distinct visual pattern on the image plane.
This pattern serves as the basis for estimating key motion parameters, specifically the rotational axis and the reference position of the center of rotation in the camera frame.
With these estimates, we can impose coplanarity and collinearity constraints, which are essential for estimating the transformation $T_{bc}$ that will be described later in Section \ref{calib}.

\subsection{Exploratory Robot Motions}
\label{param:motion}
\begin{figure}[!t]   

    \centerline{\input{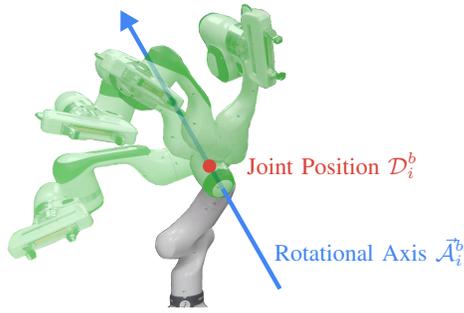}}
    \caption{Green overlay indicates parts of the robot that are considered as the rotation of a single rigid body. The motion could be parameterized with the rotation axis $\mathcal{\vec{A}}^b_i$ and the joint position $\mathcal{D}^b_i$}
    \label{pic:motion}
    \vspace{-0.5cm}
\end{figure}

Exploratory motion consists of a single joint movement, which can be regarded as the rotation of a single rigid body, as shown in Fig. \ref{pic:motion}.
For $N$ robot motions, each exploratory motion $\mathcal{U}_i$ where $ i=1,2,\dots,N$ is defined as a tuple $\mathcal{U}_i = (\mathcal{Q}_i, \mathcal{J}_i, \Delta_i)$.
Here, $\mathcal{Q}_i\in \mathbb{R} ^ \mathcal{N} $ represents the starting robot joint configuration, $\mathcal{J}_i \in \{1,2,\dots,\mathcal{N}\}$ denotes the selected joint for the rotation, and $\Delta_i \in \mathbb{R}$ is the change in the configuration of joint $\mathcal{J}_i$ during the motion. 
Given the environment's collision model, the motion $U_i$ is selected by randomly generating $Q_i$ and $\mathcal{J}_i$, while ensuring that the $\Delta_i$ of the resulting motion satisfies a minimum required value while avoiding collisions. 

Each robot motion, modeled as the rotation of a single rigid body, can be parameterized by its rotational axis $\mathcal{\vec{A}}^b_i \in \hat{\mathbb{R}}^3$ and joint position $\mathcal{D}^b_i \in \mathbb{R}^3$ in the robot frame. 
The exploratory motions are observed in the camera frame with rotational axis $\mathcal{\vec{A}}^c_i \in \hat{\mathbb{R}}^3$ and reference position $\mathcal{\vec{D}}^c_i \in \hat{\mathbb{R}}^3$  estimated from the keypoint trajectory's visual pattern in the camera image frame.
\subsection{Visual Patterns}
\label{param:pattern}

\begin{figure}[!t]  
    \centerline{
        \includegraphics[width=0.4\columnwidth]{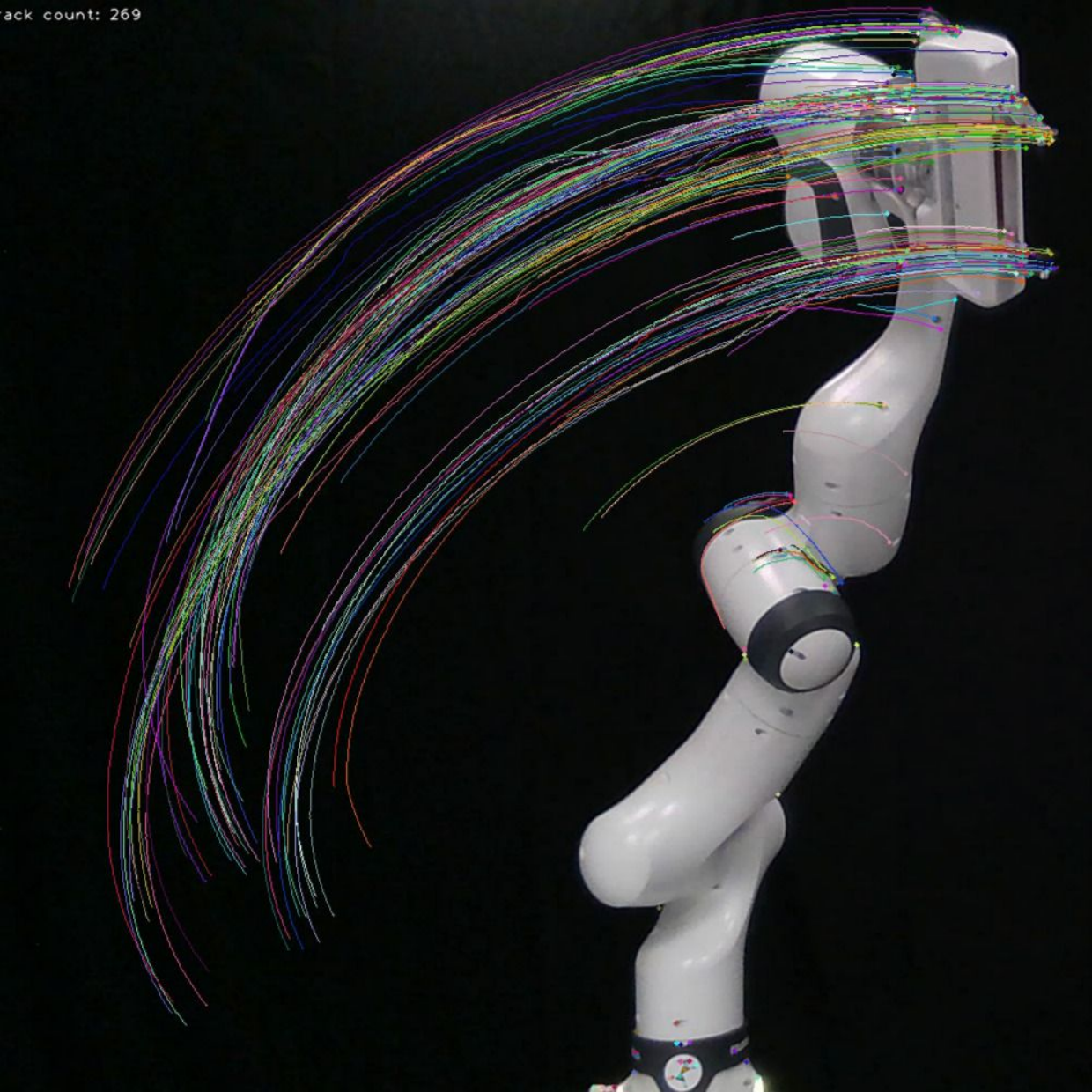}
        \hspace{5mm}        
        \input{graph/ellipse.tikz}}
    
    \caption{(left) Tracked visual features moving along 3D circles. (right) Samples of keypoint trajectories (red) and corresponding visual patterns (green) from conic functions fitting.}
     \label{pic:elpfit}
     \vspace{-0.5cm}
\end{figure}

During each motion $\mathcal{U}_i$, we track $L_i$ keypoints, which are identified by visual features such as strong corners in the image \cite{323794}.
The value of $L_i$ is dynamically determined by the tracking algorithm and may vary across different motions. 
These tracked keypoints generate trajectories that capture the robot's movement.
For each keypoint $j=1, 2,\dots, L_i$, the features are tracked across the $M_{i,j}$ consecutive frames.
This allows us to define the observation of 2D keypoint trajectories 
$\mathcal{P}^\pi_{i,j}$ as:
\begin{equation}
    \label{eq:trackpoint}
    \mathcal{P}^{\pi}_{i,j}=\{((u_{i,j,k}, v_{i,j,k}), \delta_{i,j,k})\mid k=1,2,\dots,M_{i,j}\}
\end{equation} 
where $(u_{i,j,k}, v_{i,j,k})$ are Cartesian coordinates on image plane $\pi$, and $\delta_{i,j,k}\in [0,\Delta_i]$ represents the change in configuration of joint $\mathcal{J}_i$ at the $k$-th frame. These trajectories provide a detailed representation of the robot's motion as observed by the camera, as shown in Figure \ref{pic:elpfit}.

The 3D points corresponding to the points on tracked trajectories $\mathcal{P}^\pi_{i,j}$ move along circular paths in 3D space. 
The normal vectors of the planes containing these 3D circles align with the rotational axis $\mathcal{\vec{A}}^c_i$. 
When these 3D circular trajectories are projected onto the image plane through perspective projection, the resulting keypoint trajectories $\mathcal{P}^\pi_{i,j}$ typically manifest as ellipses or straight lines.
However, distinguishing between lines caused by the robot's motion and those arising from tracking noise can be challenging. 
To address this, we focus exclusively on keypoint trajectories that exhibit elliptical shapes, as they provide more reliable information for analysis.

These ellipses can be described as a special case of a general conic function:
\begin{align}
    \label{eq:simple_ellipse}
    \Xi_{i,j}^\pi(u,v) =& A_{i,j}u^2+B_{i,j}uv+C_{i,j}v^2\\
        &+D_{i,j}u+E_{i,j}v+F_{i,j}=0 \nonumber
\end{align}
subject to the ellipse constraint:
\begin{equation}
    \label{eq:ellipse_const}
    4A_{i,j}C_{i,j}-B_{i,j}^2=0
\end{equation}
Here, $A_{i,j}$, $B_{i,j}$, $C_{i,j}$, $D_{i,j}$, $E_{i,j}$ and $F_{i,j}$ are coefficients of the ellipse on the image plane $\pi$, and $(u,v)$ represents the Cartesian coordinates of a point lying on it. 
The function $\Xi^\pi_{i,j}$ defines the algebraic distance from the point $(u,v)$ to the conic section. 

Furthermore, the 3D circles corresponding to the motion $\mathcal{U}_i$ share a common origin on the rotational axis. 
Consequently, the semi-major axes of their projected ellipses are aligned in the same direction. 
For all $L_i$ ellipses associated with the motion $\mathcal{U}_i$, let $\phi_i$ denote their orientation. Using the Cartesian representation of the ellipse, this orientation is given by $\tan 2\phi_i = \frac{B_{i,j}}{C_{i,j}-A_{i,j}}$. 
By introducing an additional parameter $G_{i,j} = C_{i,j} - A_{i,j}$, we can impose the constraints $B_{i,j} = B_i$ and $G_{i,j} = G_{i}$ for all $j = 1, 2,\dots,L_i$. 
With these constraints, Equation \eqref{eq:simple_ellipse} can be rewritten as:
\begin{align}
    \Xi^\pi_{i,j}(u,v) =& A_{i,j}(u^2+v^2)+B_i uv+G_i v^2\\
    &+D_{i,j}u+E_{i,j}v+F_{i,j}=0 \nonumber
\end{align}

To fit a set of points, i.e., the tracked keypoints, to conic functions, we adopt an approach similar to the method in \cite{oy1998NumericallySD}, minimizing the sum of squared algebraic distances from the points to their respective conics:
\begin{align}
\label{eq:og_min}
    \arg\min_{g} \sum_{j=1}^{L_i}\sum_{k=1}^{M_{i,j}} \Xi^\pi_{i,j}(u_{i,j,k}, v_{i,j,k})^2
\end{align}
which could be rearranged as 
\begin{align}
    \label{eq:ellipse_min}
    \arg\min_{g} \|Wg\|^2
\end{align}
where $g$ is a vector containing all $4 L_i+2$ coefficients and parameters of the ellipses and $W$ is a design matrix of the size $\sum_{j=1}^{L_i}M_{i,j}\times4 L_i+2$, representing the least squares minimization Equation \eqref{eq:og_min}. 

Once the solution for $g$ is obtained, we can reconstruct the conic functions $\Xi^\pi_{i,j}$ for all keypoint trajectories $j= 1,2,\dots, L_i$, ensuring that their semi-major axes share the same orientation.  
An example of this reconstruction is illustrated in Figure \ref{pic:elpfit}.

Note that the minimization problem in Equation \eqref{eq:ellipse_min} does not explicitly enforce the ellipse constraint in Equation \eqref{eq:ellipse_const}. 
As a result, solutions that do not satisfy this constraint must be discarded to ensure that only valid ellipses are retained.

\subsection{Estimating Rotational Axis}
\label{param:rotest}
\begin{figure}[!t]   

    \centerline{\input{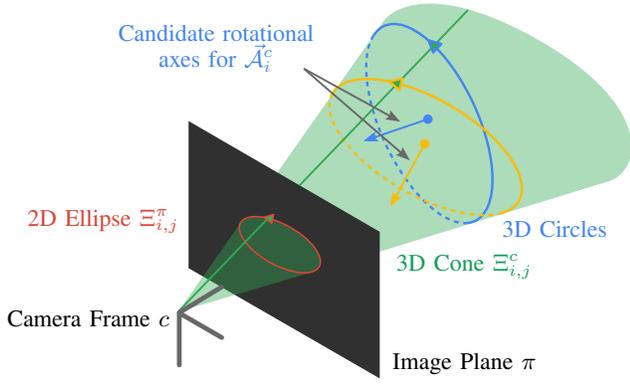}}
    \caption{A 3D cone (green) indicates the possible 3D circle candidates (blue and yellow). When projected onto the image plane, these candidates generate the visual pattern of a 2D ellipse (red)}
    \label{pic:3dcone}
    \vspace{-0.5cm}
\end{figure}

 For the rotational axis $\mathcal{\vec{A}}^b_i$ of the motion $\mathcal{U}_i$, shown in Figure \ref{pic:motion}, the corresponding rotational axis in the camera frame $\mathcal{\vec{A}}^c_i$ can be estimated using the 3D orientation (i.e., the surface normal) of the 3D circle.
 A closed-form solution for this problem has been introduced in\cite{163786}. 
 
First, we define a 3D cone surface whose base is the perspective projection of the 3D circle (represented as the ellipse $\Xi^\pi_{i,j}$) and whose vertex is the center of the camera, as illustrated in Fig. \ref{pic:3dcone}. 
Determining the 3D orientation of the circle involves finding a plane that intersects the cone and generates a circular curve.

The equation of the 3D cone, with the ellipse  $\Xi^\pi_{i,j}$ as its base, can be constructed using the ellipse's coefficients as:
\begin{align}
    \label{eq:3dcone}
    \Xi^c_{i,j}(x,y,z) =& A_{i,j}x^2+B_{i,j} xy+C_{i,j} y^2 \\
    &+D_{i,j}xz+E_{i,j}yz+F_{i,j}z^2=0 \nonumber
\end{align}
where $(x, y, z)$ are Homogeneous coordinates in the camera frame $c$ of the points lying on the cone. 
To simplify the representation, we introduce a symmetric matrix $Q_{i,j}$
\begin{align}
    Q_{i,j} &= \begin{bmatrix}
        A_{i,j} & \frac{B_{i,j}}{2} & \frac{D_{i,j}}{2}\\
        \frac{B_{i,j}}{2} & C_{i,j} & \frac{E_{i,j}}{2}\\
        \frac{D_{i,j}}{2} & \frac{E_{i,j}}{2} & F_{i,j}
    \end{bmatrix}  \nonumber \\
    \Xi^c_{i,j}(x,y,z) &= \begin{bmatrix}
    x & y & z
\end{bmatrix}Q_{i,j}\begin{bmatrix}
    x & y & z
\end{bmatrix}^T = 0
\end{align}

Let ${}^1\lambda_{i,j}, {}^2\lambda_{i,j}, {}^3\lambda_{i,j}$ denote the eigenvalues of $Q_{i,j}$, with  corrsponding eigenvectors ${}^1e_{i,j}, {}^2e_{i,j}, {}^3e_{i,j}$. 
The equation of the 3D cone can be further simplified as:
\begin{equation}
    {}^1\lambda_{i,j} X^2 + {}^2\lambda_{i,j} Y^2 +  {}^3\lambda_{i,j} Z^2 = 0
\end{equation}
where the $(X,Y,Z)$ are Homogeneous coordinates in the canonical frame of conicoids, where the principal axis of the central cone aligns with the Z axis.
Without loss of generality, we can order the eigenvalues such that ${}^3\lambda_{i,j}\!<\!0\!<\!{}^1\lambda_{i,j}\!<\!{}^2\lambda_{i,j}$.
The four possible solutions for the 3D circle plane normal (i.e., the rotational axis) are then given by:
\begin{align}
    a^c\!&=\!\pm\!\sqrt\frac{{}^1\lambda_{i,j}-{}^3\lambda_{i,j}}{{}^2\lambda_{i,j}-{}^3\lambda_{i,j}}\!\cdot\!{}^3e_{i,j}\!\pm\!\sqrt\frac{{}^2\lambda_{i,j}-{}^1\lambda_{i,j}}{{}^2\lambda_{i,j}-{}^3\lambda_{i,j}} \!\cdot\!{}^2e_{i,j}\\
    \vec{a}^c\!&=\!\frac{a^c}{\|a^c\|} \nonumber
\end{align}
These solutions represent two pairs of normals that correspond to the same 3D circle but rotate in opposite directions, as exemplified by the blue and yellow circles in Figure \ref{pic:3dcone}.  
To resolve this ambiguity, we eliminate solutions that rotate in the opposite direction of the observed keypoint trajectory $\mathcal{P}^\pi_{i,j}$, resulting in two candidate rotational axes per keypoint trajectory. 
This allows us to generate a set of candidate rotational axes for $\mathcal{\vec{A}}^c_i$ as $\mathcal{\vec{A}}'_i = \{\vec{a}^c_{i,l}\mid l=1,2,\dots, 2L_i\}$. 






\subsection{Estimating the Reference Position for the Center of Rotation}
\label{param:posest}
\begin{figure}[!t]   

    \centerline{\input{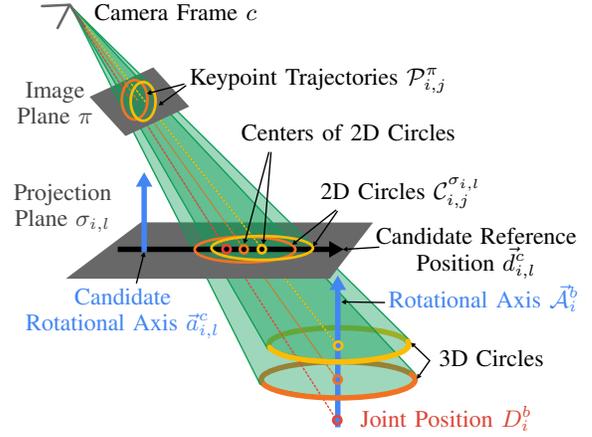}}
    \caption{Perspective projection of the keypoint trajectories $\mathcal{P}^\pi_{i,j}$ on projection plane $\sigma_{i,l}$ create the projected trajectories $\mathcal{P}^{\sigma_{i,l}}_{i,j}$. The projected trajectories create a visual pattern of 2D circles that could be used to determine the reference positions of the exploratory motion.}
    \label{pic:projected}
    \vspace{-0.5cm}
\end{figure}
For the motion $\mathcal{U}_i$ with the joint position $D_i^b$ shown in Figure \ref{pic:motion}, we can use the rotational axis candidates $\mathcal{\vec{A}}'_i$, calculated in Section \ref{param:rotest}, to estimate the corresponding reference positions $\mathcal{\vec{D}}'_i = \{\vec{d}^c_{i,l}\mid l=1,2,\dots,2L_i\}$ in the camera frame $c$. Then, we could select the best candidate from $\mathcal{\vec{A}}'_i$ and $\mathcal{\vec{D}}'_i$as estimates  $\mathcal{\vec{A}}^c_i$ and $\mathcal{\vec{D}}^c_i$ respectively.
When the trajectories $\mathcal{P}^\pi_{i,j}$ are perspective-projected onto a projection plane $\sigma_{i,l}$ with an arbitrary distance from the camera and a normal vector $\vec{a}^c_{i,l}$, the projected trajectories form 2D circles. 
Simultaneously, the 3D line representing the rotational axis projects onto the plane as a 2D line $\vec{d}^c_{i,l}$ that passes through the centers of these 2D circles, as shown in Figure. \ref{pic:projected}.
The perspective projection of $\mathcal{P}^\pi_{i,j}$ could be represented as:
\begin{equation}
    \mathcal{P}^{\sigma_{i,l}}_{i,j}\! =\! \{((u'_{i,j,k,l},v'_{i,j,k,l}),\delta_{i,j,k})\mid k\!=\!1,2,\dots,M_{i,j}\}
\end{equation}
 where $(u'_{i,j,k,l},v'_{i,j,k,l})$ are Cartesian coordinate on the projection plane $\sigma_{i,l}$. 
The 2D circle $\mathcal{C}^{\sigma_{i,l}}_{i,j}$ on the projection plane $\sigma_{i,l}$ corresponding to the robot configuration $\delta_{i,j,k}$ can be described by the polar equation:
\begin{align}
    \mathcal{C}^{\sigma_{i,l}}_{i,j}=& \{(u''_{i,j,k,l}, v''_{i,j,k,l}) \mid k=1,2,\dots,M_{i,j}\} \nonumber \\ 
    u''_{i,j,k,l} =& c^x_{i,j,l} + \alpha_{i,j,l}\sin(\delta_{i,j,k}+\beta_{i,j,l}) \\
    v''_{i,j,k,l} = & c^y_{i,j,l} + \alpha_{i,j,l}\cos(\delta_{i,j,k}+\beta_{i,j,l})) \nonumber 
\end{align} 
where $(c^x_{i,j,l}, c^y_{i,j,l})$ are the 2D coordinates of the circle's center, $\alpha_{i,j,l}$ is the radius of the circle and $\beta_{i,j,l}$ is the starting angle of the curve.  

The fitting of these 2D circles can be formulated as a minimization problem, where the goal is to minimize the sum of squared distances between the projected points $\mathcal{P}^{\sigma_{i,l}}_{i,j}$ and their corresponding circles:
\begin{align}
    \label{eq:2dcircle}
    J_{i,j,l} & = \sum_{k=1}^{M_{i,j}}\|(u'_{i,j,k,l}, v'_{i,j,k,l}) - (u''_{i,j,k,l}, v''_{i,j,k,l})\|^2 \nonumber\\
   \arg&\min_{c^x_{i,j,l}, c^y_{i,j,l}, \alpha_{i,j,l}, \beta_{i,j,l}} J_{i,j,l}
\end{align}
where $J_{i,j,l}$ is the cost function. To find the globally optimal solution to Equation \eqref{eq:2dcircle}, we consider its first-order optimality conditions:
\begin{align}
\label{eq:circlesystem}
\frac{\partial J_{i,j,l}}{\partial c^x_{i,j,l}} = 0 & & \frac{\partial J_{i,j,l}}{\partial c^y_{i,j,l}} = 0 & &\frac{\partial J_{i,j,l}}{\partial  \alpha_{i,j,l}} = 0 & & \frac{\partial J_{i,j,l}}{\partial \beta_{i,j,l}} = 0
\end{align}
This system of four equations can yield at most two critical points for $0\!\leq\!\beta_{i,j,l} \!<2\pi$. The global minimum is determined by substituting these solutions back into $J_{i,j,l}$ and selecting the one with the lower sum of errors.

The reference position of the rotation $\vec{d}^c_{i,l}$ is determined by first fitting a 2D line to the set of 2D center coordinates $(c^x_{i,j,l}, c^y_{i,j,l})$ for all $j=1,2,\dots,L_i$. The 2D line is obtained using RANSAC linear fitting to eliminate outliers trajectories. 
Once the 2D line is determined, it is converted into a 3D vector $\vec{d}^c_{i,l}$ in the camera frame. 

After determining the reference positions for all candidate rotational axes, the best candidate is selected using a heuristic function.
This function evaluates the quality of each candidate by summing the error functions from the circle-fitting process for all keypoint trajectories projected onto the plane associated with that candidate. 
To account for variations in error values caused by changes in circle size due to the projection, we normalize the error by its radius, effectively eliminating this effect.
\begin{align}
    \label{eq:costfunc}
    J'_{i,l} = \sum_{j=1}^{L_i} \frac{J_{i,j,l}}{\alpha_{i,j,l}} &&
    l^* = \arg\min_{l=1,\dots,2L_i}  J'_{i,l} 
\end{align}
where $J'_{i,l}$ is the error function of the candidate rotational axis $\vec{a}^c_{i,l}$. 
The candidate with the smallest total error is selected as the best rotational axis $\mathcal{\vec{A}}^c_i = \vec{a}^c_{i,l^*}$ with its corresponding reference position $\mathcal{\vec{D}}^c_i = \vec{d}^c_{i,l^*}$. 
As shown in Figure \ref{pic:motion}, the estimates $\mathcal{\vec{A}}^c_i$ and $\mathcal{\vec{D}}^c_i$ correspond to the rotational axis $\mathcal{\vec{A}}^b_i$ and joint position $\mathcal{D}^b_i$, respectively, as observed from the camera frame.




\section{Robot-Camera Calibration}
\label{calib}
In this section, we solve for the transformation from the camera frame to the robot frame $T_{bc}$ by imposing geometric constraints from the observation to the current robot state. 
By formulating an optimization problem that enforces collinear and coplanarity constraints introduced in \cite{agostinho2023cvxpnpl}, we ensure that the computed transformation accurately aligns the robot's movements with the camera's observations. 

\subsection{Geometric Correspondences}
\label{calib:geocon}
\begin{figure}[!t]   

    \centerline{\input{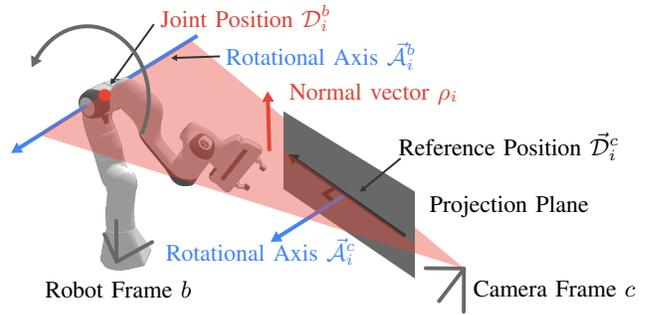}}
    \caption{\fix{Collinearity and Coplanarity constraints corresponding to the robot's exploratory motion.}}
    \label{pic:calib}
    \vspace{-0.5cm}
\end{figure}
To ensure this alignment, we formulate geometric constraints based on the collinearity of the exploratory motion's rotational axis and coplanarity of the exploratory motion's joint position, as shown in Figure \ref{pic:calib}.

For each motion $\mathcal{U}_i$, we compute the robot's current rotational axis $\mathcal{\vec{A}}_i^b$ and the position $\mathcal{D}_i^b$ of the rotating joint $\mathcal{J}_i$ using a forward kinematics function $\Gamma: (\mathbb{R}^{\mathcal{N}}, \mathbb{R}) \to (\hat{\mathbb{R}}^3, \mathbb{R}^3)$:
\begin{equation}
    \mathcal{\vec{A}}_i^b, \mathcal{D}_i^b = \Gamma(\mathcal{Q}_i, \mathcal{J}_i)
\end{equation}
These computed values, $\mathcal{\vec{A}}_i^b$ and $ \mathcal{D}_i^b$
must align with the observed rotational axis $\mathcal{\vec{A}}^c_i$ and reference position $\mathcal{\vec{D}}^c_i$ obtained from the camera frame in Section \ref{param}.

\textit{Collinearity of Rotational Axes}:
The rotational axis $\mathcal{\vec{A}}^b_i$ in the robot frame must be colinear with the observed rotational axis $\mathcal{\vec{A}}^c_i$ in the camera frame after applying the rotational transformation $R_{bc}$. This constraint can be expressed using the cross-product:
\begin{equation}
    \mathcal{\vec{A}}^c_i \times ( R_{bc}\mathcal{\vec{A}}^b_i) = 0  
\end{equation}
where $\times$ denotes the cross product. 
By representing the cross product using the skew-symmetric matrix $\lfloor \mathcal{\vec{A}}^c_i\rfloor_{\times}$ the equation can be rewritten as $\lfloor \mathcal{\vec{A}}^c_i\rfloor_{\times} R_{bc}\mathcal{\vec{A}}^b_i = 0$.
To incorporate all $N$ motions, we vectorize the matrix multiplication and rearrange terms, resulting in a linear system:
\begin{equation}
    H_r r_{bc} = 0
\end{equation}
where $H_r$ is a $3N\times9$ matrix constructed from the rotational axis constraints and $r_{bc}=vec(R_{bc})$ is a $9\times1$ vector obtained by vectorizing $R_{bc}$.

\textit{Coplanarity of Reference Points}:
The reference position $\mathcal{\vec{D}}^c_i$, when not colinear with $\mathcal{\vec{A}}^c_i$, forms a unique plane that intersects with the origin point in the camera frame. The normal of this plane is given by $\rho_i=\mathcal{\vec{D}}^c_i \times \mathcal{\vec{A}}^c_i$ where the corresponding joint position $\mathcal{D}^b_i$ must lie on this plane.
Then, We could impose a constraint using an algebraic distance from the point to the plane:
\begin{equation}
    \label{eq:trancon}
    \rho_i\cdot (R_{bc} \mathcal{D}_i^b+t_{bc}) = 0
\end{equation}
Similar to the collinearity constraint, we vectorize this equation for all $N$ motions, resulting in:
\begin{equation}
    H_p r_{bc} + K_p t_{bc} = 0
\end{equation}
where $H_p$ is a $3N\times9$ matrix, and $K_p$ is a $3N\times3$ matrix, constructed from the plane normal vectors and joint position.

These constraints form the basis of a convex optimization problem, which solves the transformation $T_{bc}$ by minimizing the discrepancies between the robot's kinematic parameters and the observed motion parameters in the camera frame.

\subsection{Convex Formulation}
Combining the two conditions (collinearity and coplanarity) from Section \ref{calib:geocon}, we construct the matrices $H$ and $K$ as follows: 
\begin{align}
    H = \begin{bmatrix}H_r & H_p\end{bmatrix}^T &,&  K = \begin{bmatrix}\pmb{0}_{3N\times3} & K_p\end{bmatrix}^T
\end{align}
\vspace{-0.1cm}
This allows us to express the combined system as
\begin{equation}
    Hr_{bc} + Kt_{bc} = 0
\end{equation}

Given the optimal solution $r^*_{bc}$ for the rotation parameters, it is known that the optimal unconstrained solution for the translation vector $t^*_{bc}$ can be computed as
\begin{equation}
    \label{eq:tranops}
    t^*_{bc} = -(K^TK)^{-1}K^THr^*_{bc}
\end{equation}
Substituting this expression back into the system, we obtain a complete system $Sr_{bc} = 0$ with:
\begin{align}
    S &= (I_{6N} - (K^TK)^{-1}K^T)H 
\end{align}
where $S$ is a $6N\times9$ matrix and $I_{6N}$ is the identity matrix of size $6N$. This formulation allows us to solve for $r_{bc}$ independently of $t_{bc}$.

We can then frame the problem as a minimization of the squared norm of the residual:
\begin{equation}
    r^*_{bc} = \arg\min_{r_{bc}} \|Sr_{bc}\|^2
\end{equation}
This optimization problem can be solved as a minimization problem similar to the ellipse fitting in Equation \ref{eq:ellipse_min}.
However, since the constraints on the rotational matrix parameters (e.g., orthogonality and unit determinant) are non-convex, they are temporarily relaxed during the optimization. 
This simplification allows us to solve the problem efficiently.
Once the optimal solution $r^*_{bc}$ is obtained, we recover the rotational matrix $R_{bc}^*$ by reshaping the vectorized form. Then, we enforce the constraints of a valid rotation matrix (orthogonality and unit determinant) by performing a singular value decomposition (SVD) on $R^*_{bc}$:
\begin{align}
R^*_{bc}\! = vec^{-1} (r^*_{bc}) && 
    U\Sigma V^T\! =\! svd(R^*_{bc})&&
    R'_{bc}\! =\! UV^T 
\end{align}
This step ensures that $R'_{bc}$ is a proper rotation matrix, satisfying all necessary geometric constraints.

Finally, we recover the optimal translation vector $t_{bc}^*$ by substituting the rotation matrix $R'_{bc}$ back into Equation \ref{eq:tranops}. With both the rotation matrix $R'_{bc}$ and the translation vector $t_{bc}^*$ determined, we obtain the complete transformation from the robot frame to the camera frame $T_{bc}$.


\vspace{-0.1cm}
\subsection{Observation Pruning and Selection}
\label{calib:select}
During the calibration process, errors in tracking keypoints can arise due to various factors such as occlusions, lighting conditions, or other external disturbances. 
These errors can lead to observations that are unsuitable for calibration, as they may introduce significant inaccuracies. 
To address this, we implement a filtering mechanism to prune out unreliable observations and retain only those that meet certain quality criteria. 
This process is divided into two scenarios: when no initial transformation estimate is available and when an initial transformation estimate exists.

\subsubsection{Filtering Observations Without an Initial Transformation Estimate}
At the start of the calibration process, with no estimated transformation $T_{bc}$, we filter observations based on the consistency of keypoint trajectories, as described in Section \ref{param}. 
Keypoint trajectories should resemble ellipses, which are projections of 3D circular motions.
Each ellipse corresponds to two possible 3D circles, and if the trajectories share a rotational axis, one solution should align with the candidate $\mathcal{\vec{A}}^c_{i}$. 
This ensures most trajectories agree with the chosen rotational axis. 
In ambiguous cases, where two candidates are very similar, we filter out observations where the difference in agreement between the top two candidates is below a certain threshold, ensuring reliability.
These strict conditions guarantee an accurate initial estimation, even without prior transformation.

\subsubsection{Filtering Observations With an Initial Transformation Estimate}
Once an initial transformation estimate $T^{i}_{bc}$ with the associated rotational matrix $R^{i}_{bc}$ and translational vector $t_{bc}^i$
is obtained from the optimization using observations from actions $\mathcal{U}_1, \dots, \mathcal{U}_{i}$, we can refine the filtering process. Note that only three observations satisfying the previous conditions are needed to compute this initial estimate. 
With $T^{i-1}_{bc}$ available, we select observations based on their reprojection errors for the rotational axis and the reference position.

The reprojection errors are calculated as follows. For the rotational axis, the error is computed as:
\begin{equation}
    \epsilon^a_i = \cos^{-1} \frac{\mathcal{\vec{A}}_i^c \cdot(R_{bc}^{i-1}\mathcal{\vec{A}}_i^b)}{\|\mathcal{\vec{A}}_i^c\|\| (R_{bc}^{i-1}\mathcal{\vec{A}}_i^b)\|}
\end{equation}
$\mathcal{\vec{A}}_i^c$ is the observed rotational axis in the camera frame, $R_{bc}^{i-1}\mathcal{\vec{A}}_i^b$ and is the reprojected rotational axis in the camera frame using the current transformation estimate. 
For the reference position, the error is computed as:
\begin{equation}
    \epsilon^d_i = |\rho_i\cdot (R_{bc}^{i-1} \mathcal{D}_i^b+t^{i-1}_{bc})|
\end{equation}
similar to Equation \ref{eq:trancon} but with the current transformation estimate.
Observations are retained only if both $\epsilon^a_i$ and $\epsilon^d_i$ are within predefined thresholds. This ensures that the selected observations are consistent with the current transformation estimate and contribute to a more accurate calibration.

\subsection{Convergence Criteria}
\label{calib:converge}
To ensure the convergence of the estimated transformation $T_{bc}$ and enable automatic termination when the algorithm is confident in the estimation, we monitor the stability of the estimates over a sliding window of $h$ robot actions.
Let $T_{bc}^{i-h:i}$ represent the sequence of estimates over the past $h$ steps, and $\gamma \in \mathbb{R}^6$ denote the dimension-wise range of these estimates.
If the range $\gamma$ remains below a predefined threshold for all six dimensions over the past $h$ steps, the calibration process terminates and returns the current estimate $T_{bc}^i$ as the final transformation from the robot to the camera frame.





\section{Experimental Evaluation}
\label{experiment}
We evaluate the proposed calibration method on the Franka Research 3 robot in both real-world and simulated environments. 
In the simulation, we use PyBullet \cite{benelot2018} to generate visual feedback for the calibration process. 
In the real-world setup, we record the robot using a RealSense Camera D415. 
In both cases, the camera resolution is set to 1920×1080 pixels, and recordings are captured at a frame rate of 30 Hz.

\subsection{Experiment in Simulation}
\begin{figure}[!t]
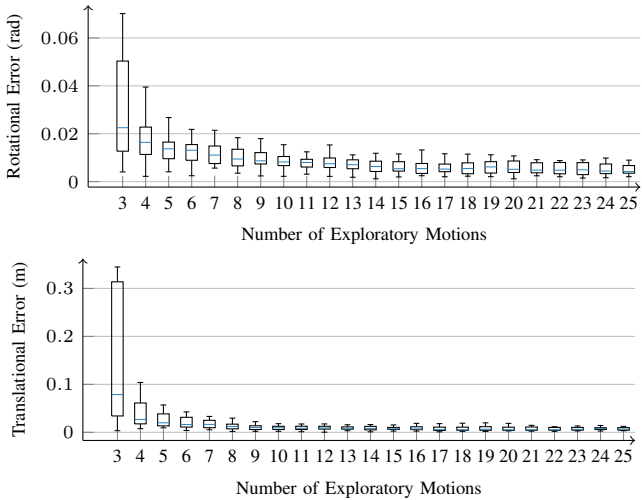
   
    \centerline{\input{graph/sim_rot.tikz}}
    \centerline{\input{graph/sim_tran.tikz}}
    \caption{(Top) Rotational Error in rad and (Bottom) Translational Error in m of the ARC-Calib method in the simulation over the number of exploratory motions. }
    \label{pic:sim_graph}
    \vspace{-0.5cm}
\end{figure}

Since there is no ground truth in real-world scenarios, we first conducted a simulation study to quantitatively evaluate our algorithm's performance under various robot-camera transformations.
The intrinsic parameters of the monocular camera were set to match those of the real-world camera. 
We perform the calibration process for 10 different camera poses, with 5 runs per setup, and calculate the calibration error using both translational and rotational metrics.
Figure \ref{pic:sim_graph} shows the accuracy of the proposed camera-to-robot calibration method. 
As expected, both rotational and translational errors decrease progressively as more observations from exploratory motions are collected.
With just 3 exploratory motions, the average rotational error is 0.0225 rad, and the translational error is 0.0786 m.
By increasing the number of exploratory motions to 25, the rotational error drops to 0.0042 rad, and the translational error reduces to 0.0065 m.
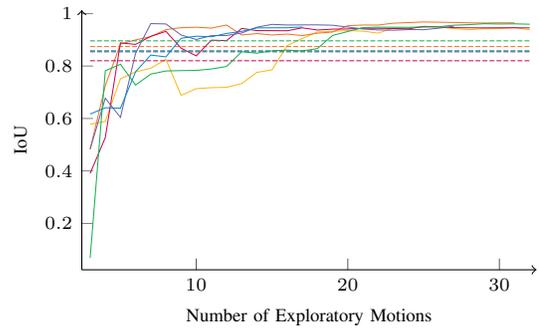
\begin{figure}[!t]   
    \centerline{
\begin{tikzpicture}

\definecolor{chocolate24311233}{RGB}{243,112,33}
\definecolor{crimson204076}{RGB}{204,0,76}
\definecolor{dimgray85}{RGB}{85,85,85}
\definecolor{dodgerblue0137208}{RGB}{0,137,208}
\definecolor{gainsboro219}{RGB}{219,219,219}
\definecolor{gainsboro229}{RGB}{229,229,229}
\definecolor{orange25218317}{RGB}{252,183,17}
\definecolor{seagreen1317775}{RGB}{13,177,75}
\definecolor{slateblue10096170}{RGB}{100,96,170}

\begin{axis}[
width=0.7\columnwidth,
height=0.4\columnwidth,
scale only axis,
axis x line*=bottom,
axis y line*=left,
tick pos=left,
xlabel={Number of Exploratory Motions},
xmin=2.45, xmax=32.45,
xtick style={color=dimgray85},
y grid style={gainsboro219},
ylabel={IoU},
ymin=0.0232259729384145, ymax=1.01300659901537,
]
\addplot [very thin, orange25218317]
table {%
3 0.578077155941658
4 0.588758682293399
5 0.751208044597759
6 0.777634471216451
7 0.791664416885205
8 0.825464833573949
9 0.688085605756156
10 0.713659788101277
11 0.71745688128894
12 0.718737191940528
13 0.732935038828085
14 0.775661906490834
15 0.785355109102685
16 0.878227351291821
17 0.90517836917152
18 0.929951149301339
19 0.932321313719359
20 0.935230025235243
21 0.933004461274942
22 0.925319386873535
23 0.944327748327251
24 0.944010667791432
25 0.950853474710067
26 0.948136272311141
27 0.942654968874792
28 0.938494619869026
29 0.941975741330514
30 0.942636229526754
31 0.946420747279409
32 0.93806178186629
};
\addplot [very thin, orange25218317, dash pattern=on 1.85pt off 0.8pt]
table {%
3 0.855470643696418
4 0.855470643696418
5 0.855470643696418
6 0.855470643696418
7 0.855470643696418
8 0.855470643696418
9 0.855470643696418
10 0.855470643696418
11 0.855470643696418
12 0.855470643696418
13 0.855470643696418
14 0.855470643696418
15 0.855470643696418
16 0.855470643696418
17 0.855470643696418
18 0.855470643696418
19 0.855470643696418
20 0.855470643696418
21 0.855470643696418
22 0.855470643696418
23 0.855470643696418
24 0.855470643696418
25 0.855470643696418
26 0.855470643696418
27 0.855470643696418
28 0.855470643696418
29 0.855470643696418
30 0.855470643696418
31 0.855470643696418
32 0.855470643696418
};
\addplot [very thin, chocolate24311233]
table {%
3 0.482577019463036
4 0.724849241069562
5 0.883837025404963
6 0.899529996036265
7 0.912048499238952
8 0.938756644019802
9 0.946005961937557
10 0.947885371613687
11 0.944519704917939
12 0.956216562940222
13 0.918347554448856
14 0.923690988219387
15 0.918263260915389
16 0.921733670907174
17 0.91611945135449
18 0.925294059572747
19 0.929487797723518
20 0.953666931872497
21 0.956270753339918
22 0.956087260771623
23 0.963273537042831
24 0.96548857276041
25 0.968016570557322
26 0.966962800596603
27 0.966022659748823
28 0.96452440276831
29 0.963833260524682
30 0.964704026256393
31 0.964517305449777
};
\addplot [very thin, chocolate24311233, dash pattern=on 1.85pt off 0.8pt]
table {%
3 0.874202352865455
4 0.874202352865455
5 0.874202352865455
6 0.874202352865455
7 0.874202352865455
8 0.874202352865455
9 0.874202352865455
10 0.874202352865455
11 0.874202352865455
12 0.874202352865455
13 0.874202352865455
14 0.874202352865455
15 0.874202352865455
16 0.874202352865455
17 0.874202352865455
18 0.874202352865455
19 0.874202352865455
20 0.874202352865455
21 0.874202352865455
22 0.874202352865455
23 0.874202352865455
24 0.874202352865455
25 0.874202352865455
26 0.874202352865455
27 0.874202352865455
28 0.874202352865455
29 0.874202352865455
30 0.874202352865455
31 0.874202352865455
32 0.874202352865455
};
\addplot [very thin, crimson204076]
table {%
3 0.390435939800723
4 0.527027562445808
5 0.888280420158149
6 0.882424973049843
7 0.913048332083645
8 0.931991030321093
9 0.868172351576533
10 0.838571947582922
11 0.898619142648058
12 0.896792810328909
13 0.943279549196715
14 0.935025081876292
15 0.934785878749303
16 0.934328574194147
17 0.945512268953984
18 0.937180733815487
19 0.940572820835543
20 0.941192457557687
21 0.943798610201746
22 0.943811311679874
23 0.942820783062805
24 0.941546557395197
25 0.950897518473061
26 0.949353480022203
27 0.946213736766734
28 0.945586280101488
29 0.945306845515925
30 0.945676761485492
31 0.945668353827762
32 0.945201647048076
};
\addplot [very thin, crimson204076, dash pattern=on 1.85pt off 0.8pt]
table {%
3 0.820078430531586
4 0.820078430531586
5 0.820078430531586
6 0.820078430531586
7 0.820078430531586
8 0.820078430531586
9 0.820078430531586
10 0.820078430531586
11 0.820078430531586
12 0.820078430531586
13 0.820078430531586
14 0.820078430531586
15 0.820078430531586
16 0.820078430531586
17 0.820078430531586
18 0.820078430531586
19 0.820078430531586
20 0.820078430531586
21 0.820078430531586
22 0.820078430531586
23 0.820078430531586
24 0.820078430531586
25 0.820078430531586
26 0.820078430531586
27 0.820078430531586
28 0.820078430531586
29 0.820078430531586
30 0.820078430531586
31 0.820078430531586
32 0.820078430531586
};
\addplot [very thin, slateblue10096170]
table {%
3 0.483172051853914
4 0.677095543091461
5 0.60428982947862
6 0.849166668670358
7 0.961541334600963
8 0.960358792370623
9 0.918177273675883
10 0.90108926238515
11 0.915956063915117
12 0.916343939683016
13 0.925914333053819
14 0.947933183905729
15 0.958653452154481
16 0.956358589806524
17 0.956651672758112
18 0.956835851485586
19 0.95535028469529
20 0.948467123317558
21 0.941070837590112
22 0.937552655814565
23 0.936382985661042
24 0.938952117311984
25 0.938427126990284
26 0.944040422209794
27 0.951586716342608
};
\addplot [very thin, slateblue10096170, dash pattern=on 1.85pt off 0.8pt]
table {%
3 0.85401713376381
4 0.85401713376381
5 0.85401713376381
6 0.85401713376381
7 0.85401713376381
8 0.85401713376381
9 0.85401713376381
10 0.85401713376381
11 0.85401713376381
12 0.85401713376381
13 0.85401713376381
14 0.85401713376381
15 0.85401713376381
16 0.85401713376381
17 0.85401713376381
18 0.85401713376381
19 0.85401713376381
20 0.85401713376381
21 0.85401713376381
22 0.85401713376381
23 0.85401713376381
24 0.85401713376381
25 0.85401713376381
26 0.85401713376381
27 0.85401713376381
28 0.85401713376381
29 0.85401713376381
30 0.85401713376381
31 0.85401713376381
32 0.85401713376381
};
\addplot [very thin, dodgerblue0137208]
table {%
3 0.616819561159205
4 0.640884466358335
5 0.639088720812641
6 0.778152402879726
7 0.841890710301653
8 0.834799821829028
9 0.905712069471498
10 0.91374307935709
11 0.911440993279704
12 0.924158025009236
13 0.932114184499213
14 0.945269036813425
15 0.945580133430412
16 0.945846310494403
17 0.948537883402532
};
\addplot [very thin, dodgerblue0137208, dash pattern=on 1.85pt off 0.8pt]
table {%
3 0.858161304975469
4 0.858161304975469
5 0.858161304975469
6 0.858161304975469
7 0.858161304975469
8 0.858161304975469
9 0.858161304975469
10 0.858161304975469
11 0.858161304975469
12 0.858161304975469
13 0.858161304975469
14 0.858161304975469
15 0.858161304975469
16 0.858161304975469
17 0.858161304975469
18 0.858161304975469
19 0.858161304975469
20 0.858161304975469
21 0.858161304975469
22 0.858161304975469
23 0.858161304975469
24 0.858161304975469
25 0.858161304975469
26 0.858161304975469
27 0.858161304975469
28 0.858161304975469
29 0.858161304975469
30 0.858161304975469
31 0.858161304975469
32 0.858161304975469
};
\addplot [very thin, seagreen1317775]
table {%
3 0.0682160013964577
4 0.781715636680097
5 0.806794129263267
6 0.727444363492726
7 0.769187381878744
8 0.781194789066394
9 0.782486043457305
10 0.783278717437198
11 0.78873436923778
12 0.798193945687002
13 0.854633654377226
14 0.849091368316488
15 0.857763353082653
16 0.860757195118086
17 0.857108590751024
18 0.866552992568306
19 0.916616195642806
20 0.931577186740229
21 0.946404985317102
22 0.947972708699278
23 0.947734271107778
24 0.947533610483312
25 0.948457122868302
26 0.949567289848218
27 0.955542137810571
28 0.957659901050398
29 0.961171323954246
30 0.96002238975895
31 0.960653574518814
32 0.959232322925909
};
\addplot [very thin, seagreen1317775, dash pattern=on 1.85pt off 0.8pt]
table {%
3 0.8959508968750364
4 0.8959508968750364
5 0.8959508968750364
6 0.8959508968750364
7 0.8959508968750364
8 0.8959508968750364
9 0.8959508968750364
10 0.8959508968750364
11 0.8959508968750364
12 0.8959508968750364
13 0.8959508968750364
14 0.8959508968750364
15 0.8959508968750364
16 0.8959508968750364
17 0.8959508968750364
18 0.8959508968750364
19 0.8959508968750364
20 0.8959508968750364
21 0.8959508968750364
22 0.8959508968750364
23 0.8959508968750364
24 0.8959508968750364
25 0.8959508968750364
26 0.8959508968750364
27 0.8959508968750364
28 0.8959508968750364
29 0.8959508968750364
30 0.8959508968750364
31 0.8959508968750364
32 0.8959508968750364
};
\end{axis}

\end{tikzpicture}}
    \caption{IoU values represent calibration error over the number of exploratory motions. Solid lines indicate the performance of ARC-Calib, while dashed lines show the traditional calibration results. Each camera setup is represented by matching colors for both methods.}
    \label{pic:real_graph}
    \vspace{-0.5cm}
\end{figure}

\begin{figure*}[th!]   

    \centerline{
        \begin{minipage}[b]{.66\columnwidth}
            \includegraphics[width=\linewidth,keepaspectratio]{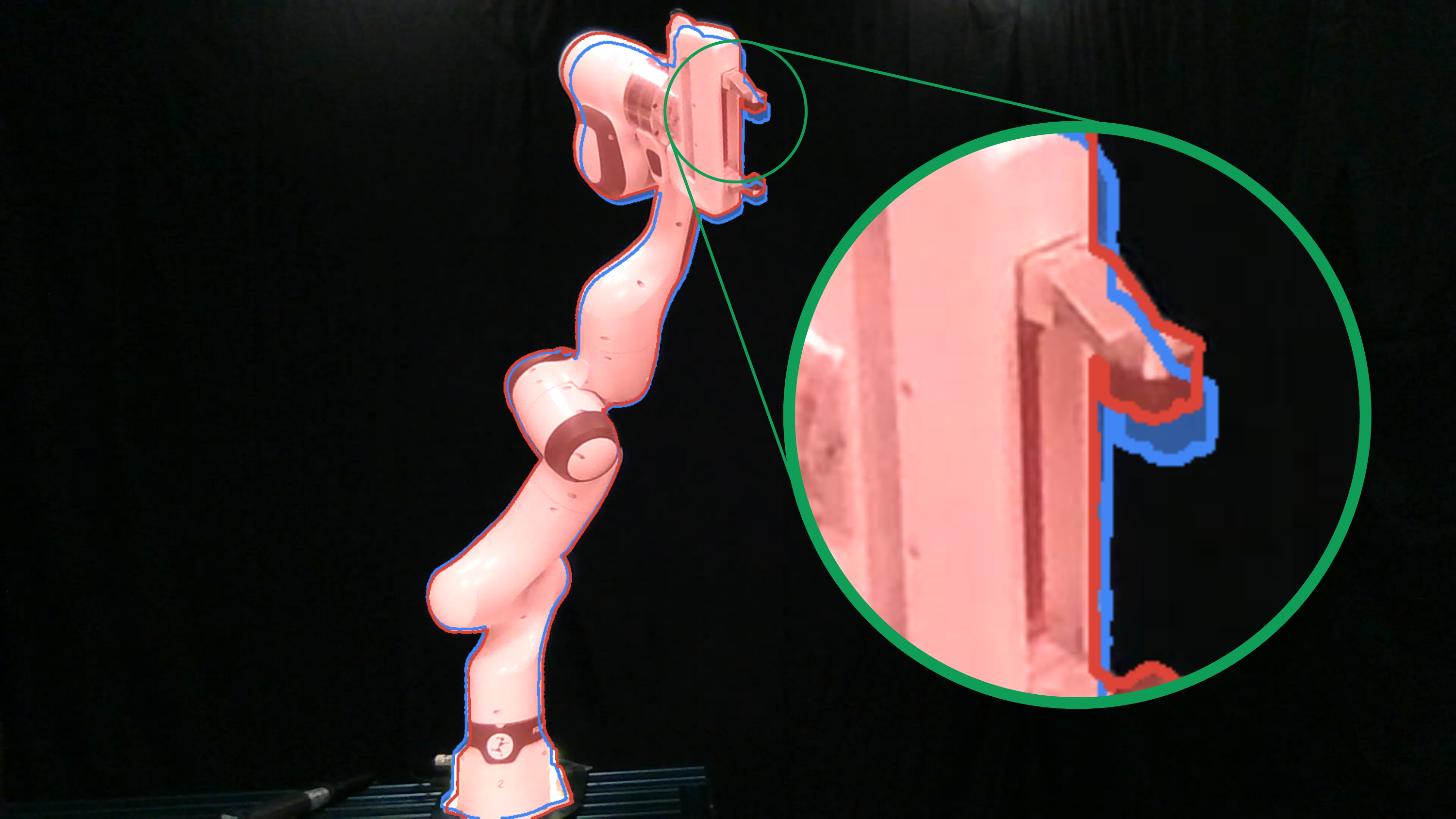}
        \end{minipage}
        \begin{minipage}[b]{.66\columnwidth}
            \includegraphics[width=\linewidth,keepaspectratio]{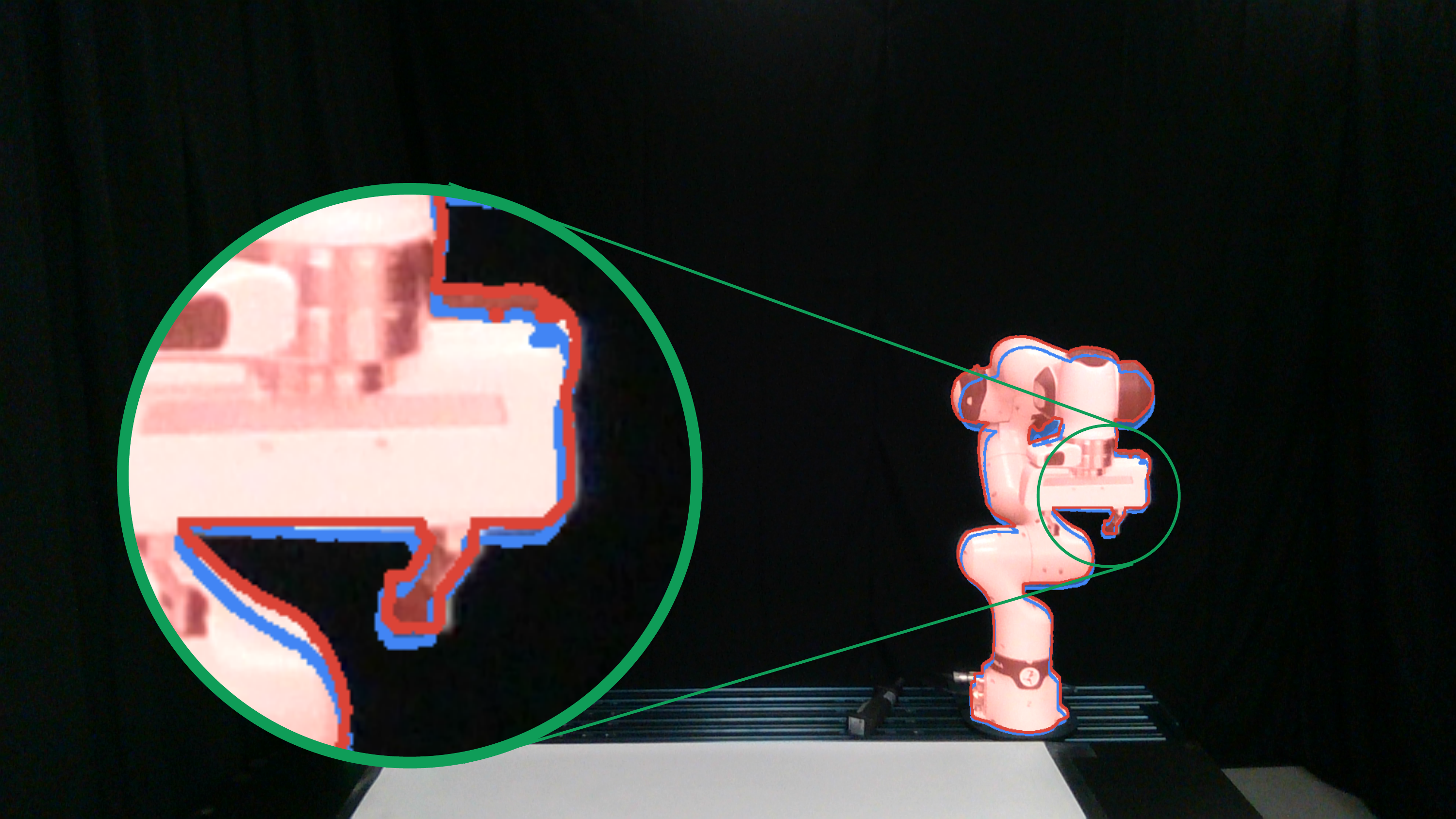}
        \end{minipage}
        \begin{minipage}[b]{.66\columnwidth}
            \includegraphics[width=\linewidth,keepaspectratio]{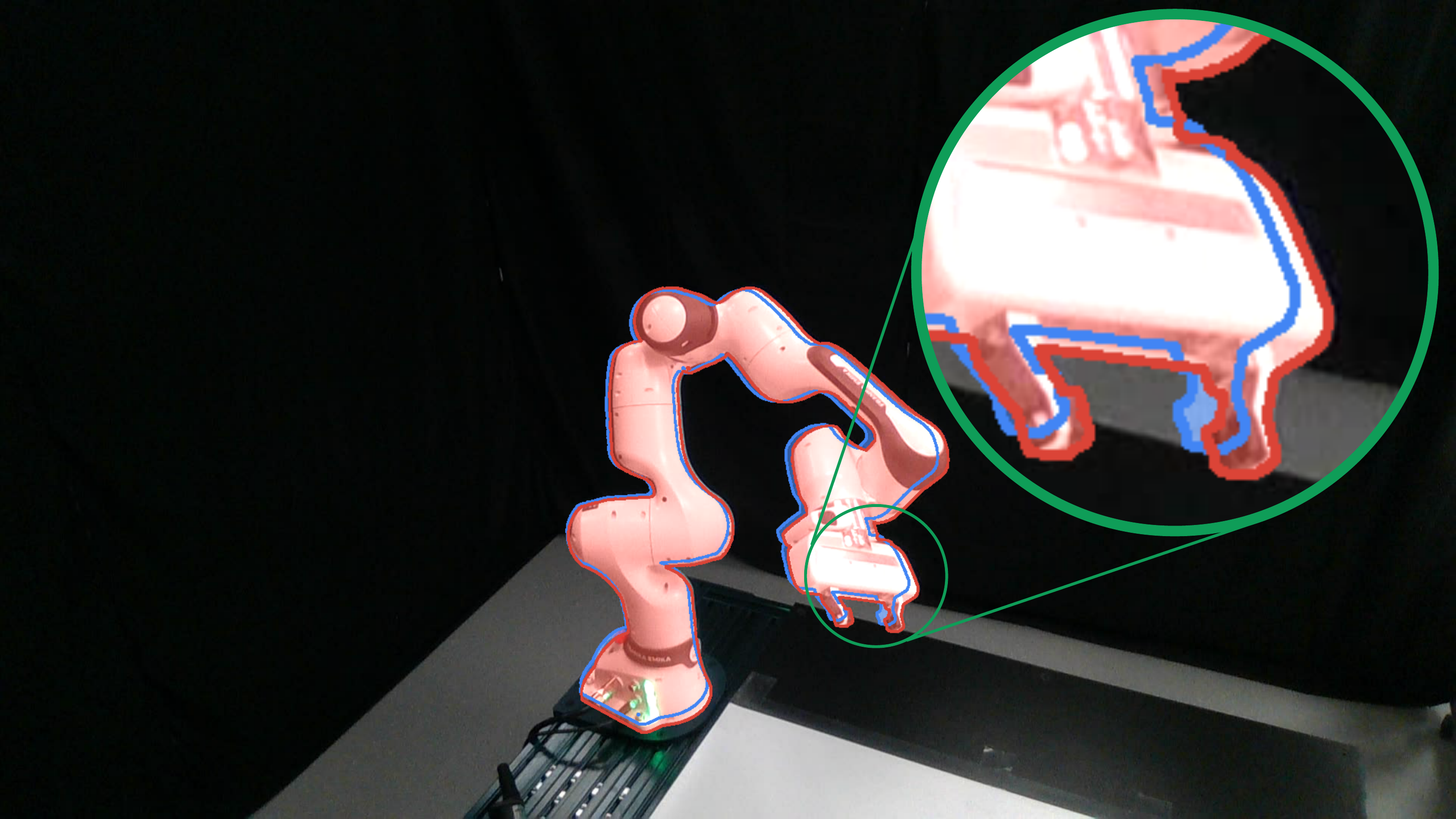}
        \end{minipage}
    }
    \vspace{0.1cm}
    \centerline{
        \begin{minipage}[b]{.66\columnwidth}
            \includegraphics[width=\linewidth,keepaspectratio]{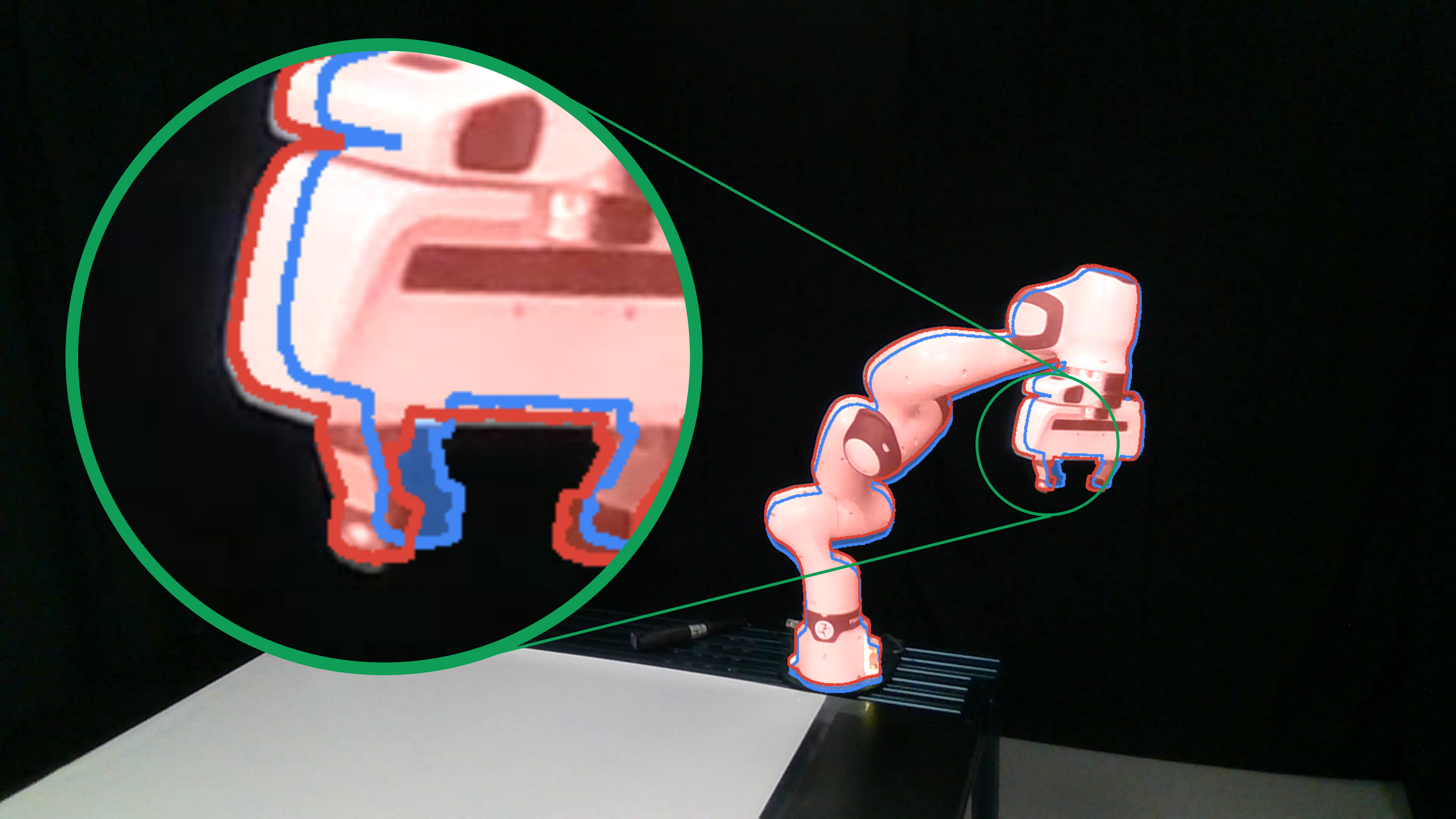}
        \end{minipage}
        \begin{minipage}[b]{.66\columnwidth}
            \includegraphics[width=\linewidth,keepaspectratio]{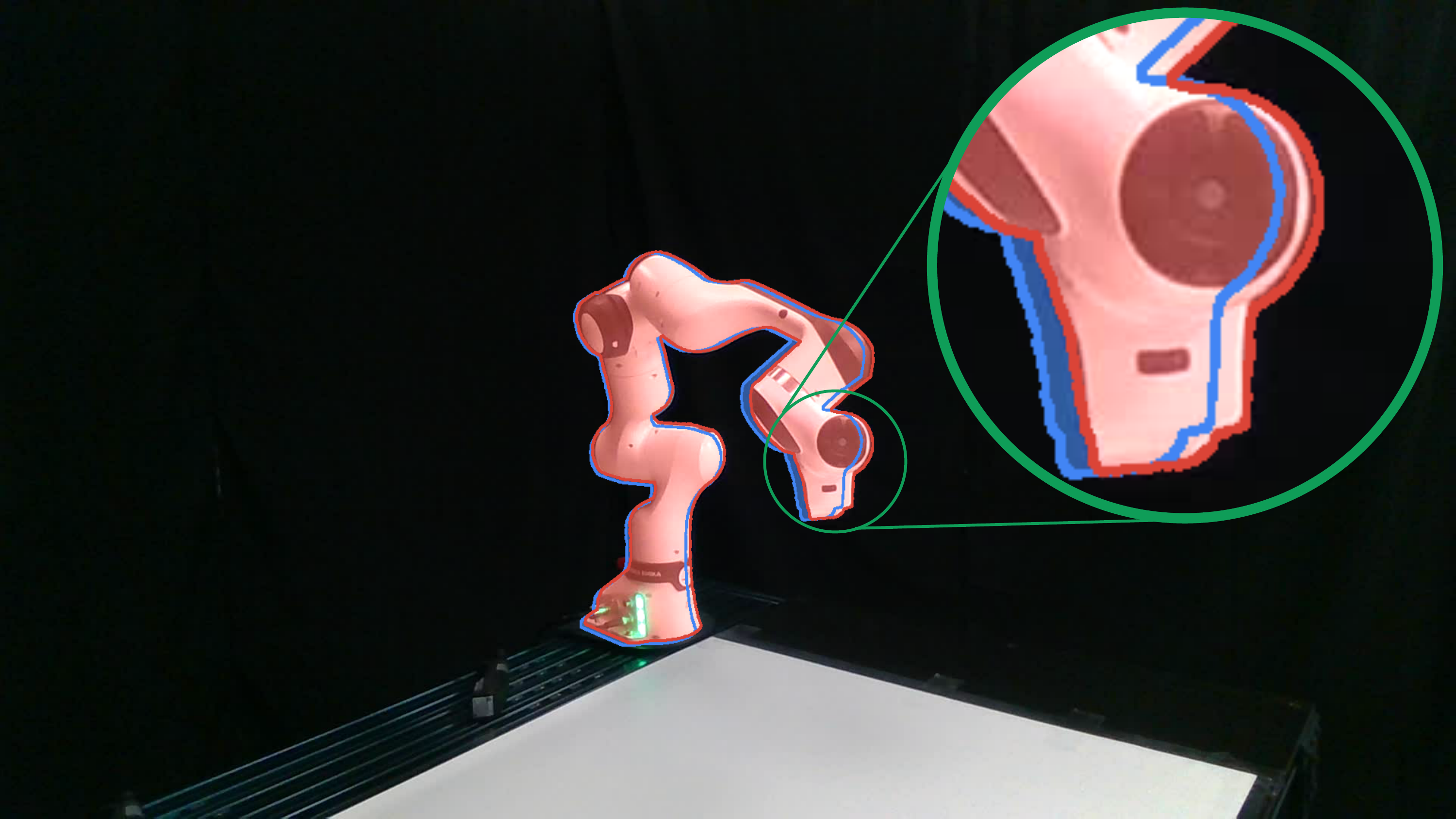}
        \end{minipage}
        \begin{minipage}[b]{.66\columnwidth}
            \includegraphics[width=\linewidth,keepaspectratio]{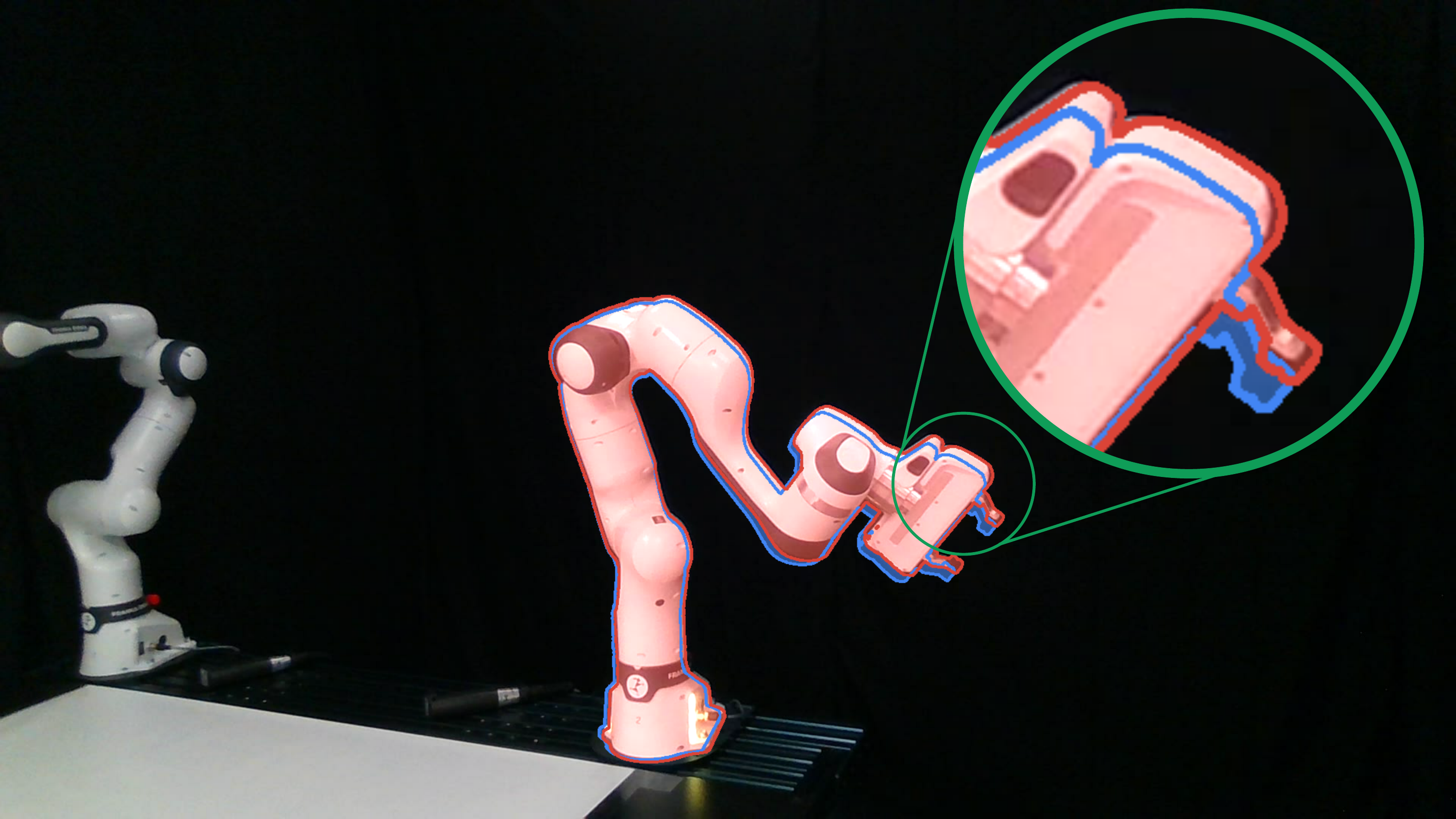}
        \end{minipage}
    }
    \caption{Qualitative results of our method on the real-world experiment with 6 different camera pose setups. The blue overlays are the robot mask rendered based on the estimation of traditional hand-eye calibration implemented in the \texttt{hand\_eye\_calibration} ROS package, and the red overlays are the robot mask rendered based on the estimation of our ARC-Calib.}
    \label{pic:real_calib}
    \vspace{-0.5cm}
\end{figure*}
\subsection{Experiment in Real-World}

We evaluate the proposed method in a real-world environment by setting up the camera in six different poses, as shown in \ref{pic:real_calib}. 
The proposed calibration method is run until the estimated results converge, as described in Section \ref{calib:converge}. 
For comparison, we run the traditional AprilTag-based hand-eye calibration method implemented in the \texttt{hand\_eye\_calibration} ROS package. 
The traditional method continuously collects data points until the calibration results converge, with the convergence point manually determined by the operator.

Since ground truth data is unavailable in the real world, the robot's projection can be used as an indirect measure of calibration accuracy. 
We render the Franka robot's mask onto the camera frame using the estimated transformations from different methods and compared them with manually labeled ground truth masks at random robot configurations. 
The Intersection over Union (IoU) metric is calculated between the projected mask and the ground truth, where a higher IoU value corresponds to lower calibration error and better accuracy.

The results show that our proposed method achieves an average IoU of 0.94, outperforming the traditional method with an average IoU of 0.84. 
Furthermore, the calibration process converges after an average of 26.5 selected robot motions. 
As shown in Figure \ref{pic:real_graph}, the trending of the calibration error using IoU in the real world is similar to that of translational/rotational error in the simulation. 

\section{Conclusion}
\label{conclusion}
In this work, we present a novel framework, ARC-Calib, for autonomous markerless camera-to-robot calibration, eliminating the need for human intervention or manual setup. 
By introducing an exploratory robot motion, the robot generates an associated visual feedback pattern that can be extracted using lightweight methods such as optical flow, replacing traditional visual markers. 
ARC-Calib is designed to be generalizable and able to operate across diverse robots and scenarios without requiring prior database knowledge, pre-trained models, or fine-tuning.
Evaluation in both physical and simulated settings demonstrates the method’s accuracy. 
This makes it a highly adaptable and efficient solution for real-world robot-to-camera calibration for vision-based robot manipulation systems.

\bibliographystyle{IEEEtran}
\bibliography{reference}

\begin{thebibliography}{10}
\providecommand{\url}[1]{#1}
\csname url@samestyle\endcsname
\providecommand{\newblock}{\relax}
\providecommand{\bibinfo}[2]{#2}
\providecommand{\BIBentrySTDinterwordspacing}{\spaceskip=0pt\relax}
\providecommand{\BIBentryALTinterwordstretchfactor}{4}
\providecommand{\BIBentryALTinterwordspacing}{\spaceskip=\fontdimen2\font plus
\BIBentryALTinterwordstretchfactor\fontdimen3\font minus \fontdimen4\font\relax}
\providecommand{\BIBforeignlanguage}[2]{{%
\expandafter\ifx\csname l@#1\endcsname\relax
\typeout{** WARNING: IEEEtran.bst: No hyphenation pattern has been}%
\typeout{** loaded for the language `#1'. Using the pattern for}%
\typeout{** the default language instead.}%
\else
\language=\csname l@#1\endcsname
\fi
#2}}
\providecommand{\BIBdecl}{\relax}
\BIBdecl

\bibitem{horaud1995hand}
R.~Horaud and F.~Dornaika, ``Hand-eye calibration,'' \emph{The International Journal of Robotics Research}, vol.~14, no.~3, pp. 195--210, 1995.

\bibitem{park1994robot}
F.~C. Park and B.~J. Martin, ``Robot sensor calibration: solving ax= xb on the euclidean group,'' \emph{IEEE Transactions on Robotics and Automation}, vol.~10, no.~5, pp. 717--721, 1994.

\bibitem{ali2019methods}
I.~Ali, O.~Suominen, A.~Gotchev, and E.~R. Morales, ``Methods for simultaneous robot-world-hand--eye calibration: A comparative study,'' \emph{Sensors}, vol.~19, no.~12, p. 2837, 2019.

\bibitem{fassi2005hand}
I.~Fassi and G.~Legnani, ``Hand to sensor calibration: A geometrical interpretation of the matrix equation ax= xb,'' \emph{Journal of Robotic Systems}, vol.~22, no.~9, pp. 497--506, 2005.

\bibitem{olson2011apriltag}
E.~Olson, ``Apriltag: A robust and flexible visual fiducial system,'' in \emph{IEEE International Conference on Robotics and Automation (ICRA)}.\hskip 1em plus 0.5em minus 0.4em\relax IEEE, 2011, pp. 3400--3407.

\bibitem{fiala2005artag}
M.~Fiala, ``Artag, a fiducial marker system using digital techniques,'' in \emph{IEEE Computer Society Conference on Computer Vision and Pattern Recognition (CVPR)}, vol.~2.\hskip 1em plus 0.5em minus 0.4em\relax IEEE, 2005, pp. 590--596.

\bibitem{garrido2014automatic}
S.~Garrido-Jurado, R.~Mu{\~n}oz-Salinas, F.~J. Madrid-Cuevas, and M.~J. Mar{\'\i}n-Jim{\'e}nez, ``Automatic generation and detection of highly reliable fiducial markers under occlusion,'' \emph{Pattern Recognition}, vol.~47, no.~6, pp. 2280--2292, 2014.

\bibitem{bennett2014chess}
S.~Bennett and J.~Lasenby, ``Chess--quick and robust detection of chess-board features,'' \emph{Computer Vision and Image Understanding}, vol. 118, pp. 197--210, 2014.

\bibitem{lu2024ctrnet}
J.~Lu, Z.~Liang, T.~Xie, F.~Ritcher, S.~Lin, S.~Liu, and M.~C. Yip, ``Ctrnet-x: Camera-to-robot pose estimation in real-world conditions using a single camera,'' \emph{arXiv preprint arXiv:2409.10441}, 2024.

\bibitem{lee2020camera}
T.~E. Lee, J.~Tremblay, T.~To, J.~Cheng, T.~Mosier, O.~Kroemer, D.~Fox, and S.~Birchfield, ``Camera-to-robot pose estimation from a single image,'' in \emph{IEEE International Conference on Robotics and Automation (ICRA)}.\hskip 1em plus 0.5em minus 0.4em\relax IEEE, 2020, pp. 9426--9432.

\bibitem{lu2023markerless}
J.~Lu, F.~Richter, and M.~C. Yip, ``Markerless camera-to-robot pose estimation via self-supervised sim-to-real transfer,'' in \emph{Proceedings of the IEEE/CVF Conference on Computer Vision and Pattern Recognition}, 2023, pp. 21\,296--21\,306.

\bibitem{labbe2021single}
Y.~Labb{\'e}, J.~Carpentier, M.~Aubry, and J.~Sivic, ``Single-view robot pose and joint angle estimation via render \& compare,'' in \emph{Proceedings of the IEEE/CVF Conference on Computer Vision and Pattern Recognition}, 2021, pp. 1654--1663.

\bibitem{horn1981determining}
B.~K. Horn and B.~G. Schunck, ``Determining optical flow,'' \emph{Artificial intelligence}, vol.~17, no. 1-3, pp. 185--203, 1981.

\bibitem{yang2018robotic}
L.~Yang, Q.~Cao, M.~Lin, H.~Zhang, and Z.~Ma, ``Robotic hand-eye calibration with depth camera: A sphere model approach,'' in \emph{International Conference on Control, Automation and Robotics (ICCAR)}.\hskip 1em plus 0.5em minus 0.4em\relax IEEE, 2018, pp. 104--110.

\bibitem{tsai1989new}
R.~Y. Tsai, R.~K. Lenz \emph{et~al.}, ``A new technique for fully autonomous and efficient 3 d robotics hand/eye calibration,'' \emph{IEEE Transactions on Robotics and Automation}, vol.~5, no.~3, pp. 345--358, 1989.

\bibitem{traslosheros2011method}
A.~Traslosheros, J.~M. Sebasti{\'a}n, E.~Castillo, F.~Roberti, and R.~Carelli, ``A method for kinematic calibration of a parallel robot by using one camera in hand and a spherical object,'' in \emph{International Conference on Advanced Robotics (ICAR)}.\hskip 1em plus 0.5em minus 0.4em\relax IEEE, 2011, pp. 75--81.

\bibitem{staranowicz2015practical}
A.~N. Staranowicz, G.~R. Brown, F.~Morbidi, and G.-L. Mariottini, ``Practical and accurate calibration of rgb-d cameras using spheres,'' \emph{Computer Vision and Image Understanding}, vol. 137, pp. 102--114, 2015.

\bibitem{lambrecht2019towards}
J.~Lambrecht and L.~K{\"a}stner, ``Towards the usage of synthetic data for marker-less pose estimation of articulated robots in rgb images,'' in \emph{International Conference on Advanced Robotics (ICAR)}, 2019, pp. 240--247.

\bibitem{zuo2019craves}
Y.~Zuo, W.~Qiu, L.~Xie, F.~Zhong, Y.~Wang, and A.~L. Yuille, ``Craves: Controlling robotic arm with a vision-based economic system,'' in \emph{Proceedings of the IEEE/CVF Conference on Computer Vision and Pattern Recognition}, 2019, pp. 4214--4223.

\bibitem{chen2023easyhec}
L.~Chen, Y.~Qin, X.~Zhou, and H.~Su, ``Easyhec: Accurate and automatic hand-eye calibration via differentiable rendering and space exploration,'' \emph{IEEE Robotics and Automation Letters}, 2023.

\bibitem{323794}
J.~Shi and Tomasi, ``Good features to track,'' in \emph{1994 Proceedings of IEEE Conference on Computer Vision and Pattern Recognition}, 1994, pp. 593--600.

\bibitem{oy1998NumericallySD}
R.~H. oy and J.~Flusser, ``Numerically stable direct least squares fitting of ellipses,'' 1998.

\bibitem{163786}
R.~Safaee-Rad, I.~Tchoukanov, K.~Smith, and B.~Benhabib, ``Three-dimensional location estimation of circular features for machine vision,'' \emph{IEEE Transactions on Robotics and Automation}, vol.~8, no.~5, pp. 624--640, 1992.

\bibitem{agostinho2023cvxpnpl}
S.~Agostinho, J.~Gomes, and A.~Del~Bue, ``Cvxpnpl: A unified convex solution to the absolute pose estimation problem from point and line correspondences,'' \emph{Journal of Mathematical Imaging and Vision}, vol.~65, no.~3, pp. 492--512, 2023.

\bibitem{benelot2018}
B.~Ellenberger, ``Pybullet gymperium,'' \url{https://github.com/benelot/pybullet-gym}, 2018--2019.

\end{thebibliography}

\end{document}